\documentclass[letterpaper, 10 pt, conference]{ieeeconf}  

\IEEEoverridecommandlockouts       

\usepackage[backend=biber,
            url=false,
            isbn=false,
            doi=false,
            backref=false,
            style=ieee,
            natbib=true,
            mincitenames=1,
            maxcitenames=1,
            citestyle=numeric-comp,
            sorting=nyt,
            block=none]{biblatex}

\DeclareUnicodeCharacter{2212}{-}

\renewcommand{\bibfont}{\small}

\usepackage{graphics}
\usepackage{bbm}
\usepackage{dsfont}
\usepackage[pdftex]{graphicx}
\usepackage{wrapfig}
\DeclareGraphicsExtensions{.pdf,.png,.jpg}
\usepackage{epsfig}
\usepackage[font={small}]{caption}
\usepackage{subcaption}
\usepackage[rightcaption]{sidecap}
\usepackage{pbox}

\usepackage{bigstrut}
\usepackage{mathtools}
\usepackage{amsmath, amssymb, amscd}
\usepackage{ wasysym } 
\usepackage{mathptmx} 
\usepackage{gensymb} 
\usepackage{nicefrac}       
\numberwithin{equation}{section} 
\usepackage{dsfont}

\DeclareMathAlphabet{\mathcal}{OMS}{lmsy}{m}{n}
\usepackage{textcomp} 

\usepackage{array} 
\usepackage{tabularx}
\usepackage{multirow}
\usepackage{multicol}
\usepackage{booktabs}
\usepackage{tabulary}

\usepackage[utf8]{inputenc}
\usepackage[english]{babel} 
\usepackage{units}
\usepackage{bm}
\usepackage{xspace}
\usepackage{flushend}
\usepackage{balance} 
\usepackage{csquotes}
\usepackage{makeidx}
\usepackage{blindtext}

\let\labelindent\relax
\usepackage{enumitem}

\usepackage{ragged2e}
\usepackage{soul} 
\usepackage{subfiles} 

\usepackage{url}
\makeatletter
\g@addto@macro{\UrlBreaks}{\UrlOrds}
\makeatother
\usepackage{color}
\usepackage[usenames,dvipsnames,table,xcdraw]{xcolor}
\usepackage{pgfplots} 
\pgfplotsset{compat=newest}

\usepackage{marginnote}
\usepackage{soul} 

\usepackage[colorinlistoftodos]{todonotes}
\newcommand{\tocite}[1]{%
\textcolor{red}{[cite:\ifthenelse{\equal{#1}{}}{}{#1}?]}
}

\newcommand{\ignore}[1]{}

\renewcommand{\baselinestretch}{1.0}



\usepackage[pdfborder={0 0 0.5}]{hyperref}
\hypersetup{
    colorlinks=true,
    linkcolor=black,
    citecolor=black,
    filecolor=cyan,
    urlcolor=black
}

\usepackage{algorithm}
\usepackage[noend]{algpseudocode}

\usepackage[switch]{lineno}
\usepackage{xspace}
\newcommand{\algname}{LazyDAgger\xspace}
\newcommand{\algabbr}{LazyDAgger\xspace}

\newcommand{\ffilt} {f}
\newcommand{\fstarfilt} {f^*}
\newcommand{\cloningloss} {\mathcal{L}(\pi_{R}(s_t),\pi_{H}(s_t))}

\newcommand{\tausup} {\beta_{H}} 
\newcommand{\tauauto} {\beta_{R}} 
\newcommand{\pisup} {\pi_{H}} 
\newcommand{\pirob} {\pi_{R}}
\newcommand{\pimeta} {\pi}
\newcommand{\indicatorint} {m_I} 
\newcommand{\indicatorsup} {m_H} 
\newcommand{\dsafe} {\mathcal{D}_{\rm S}} 
\newcommand{\lsafe} {\mathcal{L}_{\rm S}}

\DeclareMathOperator*{\argmin}{arg\,min}

\title{\LARGE \bf
\algname: Reducing Context Switching\\ in Interactive Imitation Learning
}

\author{
Ryan Hoque, Ashwin Balakrishna, Carl Putterman, Michael Luo, \\Daniel S. Brown, Daniel Seita, Brijen Thananjeyan, Ellen Novoseller, Ken Goldberg
\thanks{AUTOLAB at the University of California, Berkeley}
\thanks{Correspondence to {\tt\small ryanhoque@berkeley.edu}}
}

\begin{document}


\maketitle

\begin{abstract}
Corrective interventions while a robot is learning to automate a task provide an intuitive method for a human supervisor to assist the robot and convey information about desired behavior. However, these interventions can impose significant burden on a human supervisor, as each intervention interrupts other work the human is doing, incurs latency with each context switch between supervisor and autonomous control, and requires time to perform. We present \algname, which extends the interactive imitation learning (IL) algorithm SafeDAgger to reduce context switches between supervisor and autonomous control.
We find that \algname improves the performance and robustness of the learned policy during both learning and execution while limiting burden on the supervisor. Simulation experiments suggest that \algabbr can reduce context switches by an average of 60\% over SafeDAgger on 3 continuous control tasks while maintaining state-of-the-art policy performance. In physical fabric manipulation experiments with an ABB YuMi robot, \algname reduces context switches by 60\% while achieving a 60\% higher success rate than SafeDAgger at execution time. 

\end{abstract}

\section{Introduction}
\label{sec:introduction}
Imitation learning allows a robot to learn from human feedback and examples~\cite{argall2009survey,arora2018survey,osa2018algorithmic}.
In particular, \textit{interactive imitation learning} (IL)~\cite{EIL,hg_dagger,safe_dagger}, in which a human supervisor periodically takes control of the robotic system during policy learning, has emerged as a popular imitation learning method, as interventions are a particularly intuitive form of human feedback~\cite{EIL}. However, a key challenge in interactive imitation learning is to reduce the burden that interventions place on the human supervisor~\cite{safe_dagger, hg_dagger}.

\begin{figure}[htb!]
\center
\includegraphics[width=0.49\textwidth]{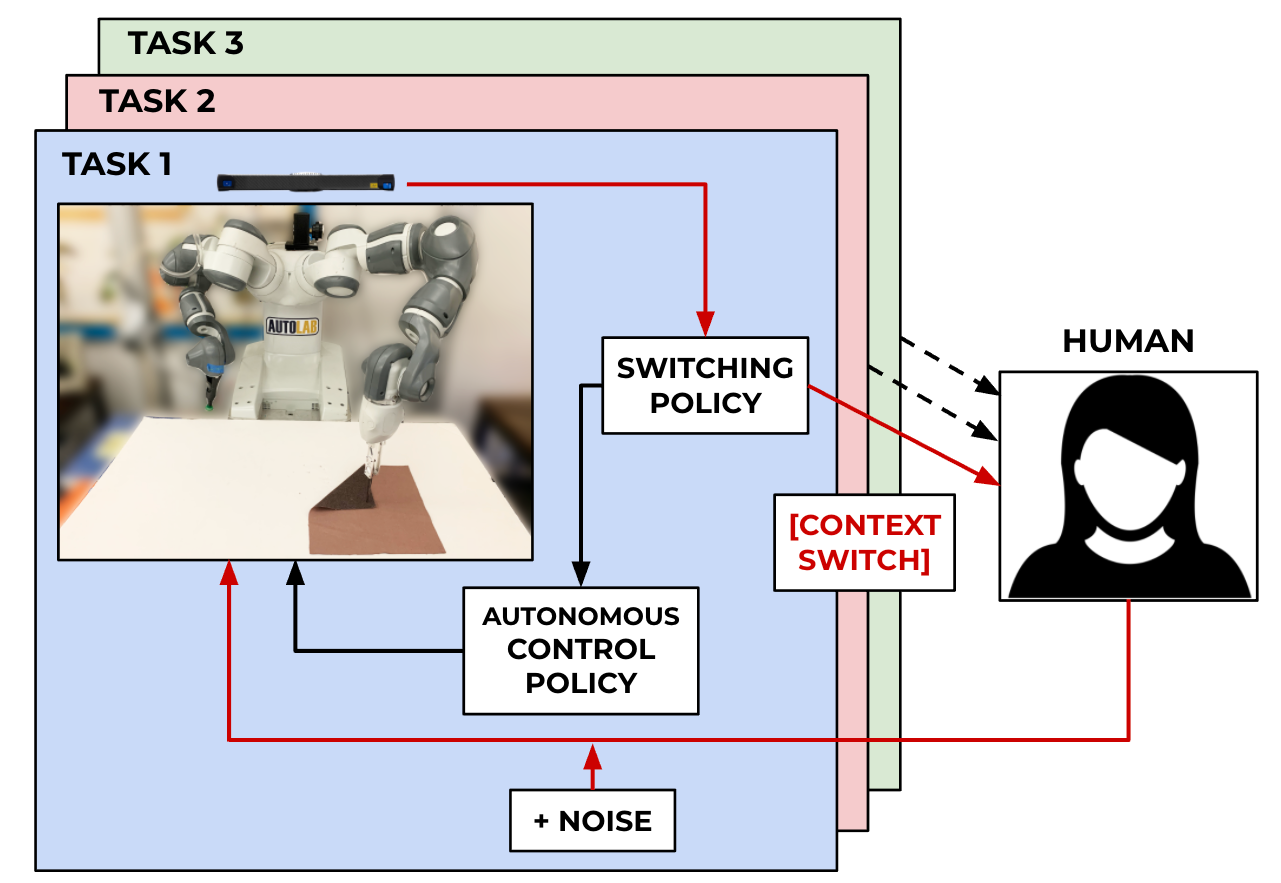}
\caption{
\algname learns to cede control to a supervisor in states in which it estimates that its actions will significantly deviate from those of the supervisor. \algname reduces context switches between supervisor and autonomous control to reduce burden on a human supervisor working on multiple tasks.}
\label{fig:teaser}
\end{figure}

One source of this burden is the cost of \textit{context switches} between human and robot control. 
Context switches incur significant time cost, as a human must interrupt the task they are currently performing, acquire control of the robot, and gain sufficient situational awareness before beginning the intervention. As an illustrative example, consider a robot performing a task for which an action takes 1 time unit and an intervention requires two context switches (one at the start and one at the end). We define \emph{latency} $L$ as the number of time units associated with a single context switch. For instance, $L \gg 1$ for a human supervisor who will need to pause an ongoing task and walk over to a robot that requires assistance. If the supervisor takes control 10 times for 2 actions each, she spends $20L + 20$ time units helping the robot. In contrast, if the human takes control 2 times for 10 actions each, she spends only $4L + 20$ time units. The latter significantly reduces the burden on the supervisor. Furthermore, prior work suggests that frequent context switches can make it difficult for the supervisor to perform other tasks in parallel~\cite{swamy2020scaled} or gain enough situational awareness to provide useful interventions~\cite{think_going}.

We present \algname (Figure~\ref{fig:teaser}), an algorithm which initiates useful interventions while limiting context switches.
The name \algname is inspired by the concept of lazy evaluation in programming language theory~\cite{lazy-eval}, where expressions are evaluated only when required to reduce computational burden. As in SafeDAgger~\cite{safe_dagger}, \algname learns a meta-controller which determines when to context switch based on the estimated discrepancy between the learner and supervisor. However, unlike SafeDAgger, \algname reduces context switching by (1) introducing asymmetric switching criteria and (2) injecting noise into the supervisor control actions to widen the distribution of visited states. One appealing property of this improved meta-controller is that even after training, \algname can be applied at execution time to improve the safety and reliability of autonomous policies with minimal context switching.
We find that across 3 continuous control tasks in simulation, \algname achieves task performance on par with DAgger~\cite{dagger} with 88\% fewer supervisor actions than DAgger and 60\% fewer context switches than SafeDAgger. In physical fabric manipulation experiments, we observe similar results, and find that at execution time, \algname achieves 60\% better task performance than SafeDAgger with 60\% fewer context switches. 

\section{Background and Related Work}
\label{sec:related-work}
Challenges in learning efficiency and reward function specification have inspired significant interest in algorithms that can leverage supervisor demonstrations and feedback for policy learning.

\textbf{Learning from Offline Demonstrations: }
Learning from demonstrations~\cite{argall2009survey,osa2018algorithmic,arora2018survey} is a popular imitation learning approach, as it requires minimal supervisor burden: the supervisor provides a batch of offline demonstrations and gives no further input during policy learning. Many methods use demonstrations directly for policy learning~\cite{pomerleau1991efficient,ijspeert2013dynamical,paraschos2013probabilistic,torabi2018behavioral}, while others use reinforcement learning to train a policy using a reward function inferred from demonstrations \cite{abbeel2004apprenticeship,ziebart2008maximum,ho2016generative,brown2019drex,airl}. Recent work has augmented demonstrations with additional offline information such as pairwise preferences~\cite{browngoo2019trex,brown2020safe}, human gaze~\cite{gaze_saran}, and natural language descriptions~\cite{tung2018reward}. While offline demonstrations are often simple to provide, the lack of online feedback makes it difficult to address specific bottlenecks in the learning process or errors in the resulting policy due to covariate shift~\cite{dagger}.



\textbf{Learning from Online Feedback: }
Many policy learning algorithms' poor performance stems from a lack of online supervisor guidance, motivating active learning methods such as DAgger, which queries the supervisor for an action in every state that the learner visits~\cite{dagger}. While DAgger has a number of desirable theoretical properties, labeling every state is costly in human time and can be a non-intuitive form of human feedback~\cite{DART}. More generally, the idea of learning from action advice has been widely explored in imitation learning algorithms~\cite{converging-supervisors,SHIV,judah2011active,jauhri2020interactive}.
There has also been significant recent interest in active preference queries for learning reward functions from pairwise preferences over demonstrations~\cite{sadigh2017active,christiano2017deep,ibarz2018reward,palan2019learning,biyik2019asking,brown2020safe}. However, many forms of human advice can be unintuitive, since the learner may visit states that are significantly far from those the human supervisor would visit, making it difficult for humans to judge what correct behavior looks like without interacting with the environment themselves~\cite{EIL,reddy2018shared}.

\textbf{Learning from Supervisor Interventions: } There has been significant prior work on algorithms for learning policies from interventions. \citet{training_wheels, ac-teach} leverage interventions from suboptimal supervisors to accelerate policy learning, but assume that the supervisors are algorithmic and thus can be queried cheaply. \citet{recovery-rl, advantage_interventions, trial_no_error} also leverage interventions from algorithmic policies, but for constraint satisfaction during learning. \citet{hg_dagger, EIL, IARL, LAND, HITL, teaching-strats} instead consider learning from human supervisors and present learning algorithms which utilize the timing and nature of human interventions to update the learned policy. By giving the human control for multiple timesteps in a row, these algorithms show improvements over methods that only hand over control on a state-by-state basis~\cite{perf-eval-IL}. However, the above algorithms assume that the human is continuously monitoring the system to determine when to intervene, which may not be practical in large-scale systems or continuous learning settings~\cite{crandall2005validating,chen2014human,swamy2020scaled,kessler2019active}. Such algorithms also assume that the human knows when to cede control to the robot, which requires guessing how the robot will behave in the future. \citet{safe_dagger} and \citet{ensemble_dagger} present imitation learning algorithms SafeDAgger and EnsembleDAgger, respectively, to address these issues by learning to request interventions from a supervisor based on measures such as state novelty or estimated discrepancy between the learner and supervisor actions. These methods can still be sample inefficient, and, as we discuss later, often result in significant context switching.
By contrast, \algabbr encourages interventions that are both easier to provide and more informative. To do this, \algname prioritizes (1) sustained interventions, which allow the supervisor to act over a small number of contiguous sequences of states rather than a large number of disconnected intervals, and (2) interventions which demonstrate supervisor actions in novel states to increase robustness to covariate shift in the learned policy.
\section{Problem Statement}
\label{sec:prob-statement}
We consider a setting in which a human supervisor is training a robot to reliably perform a task. The robot may query the human for assistance, upon which the supervisor takes control and teleoperates the robot until the system determines that it no longer needs assistance. We assume that the robot and human policy have the same action space, and that it is possible to pause task execution while waiting to transfer control. We formalize these ideas in the context of prior imitation learning literature.

We model the environment as a discrete-time Markov decision process (MDP) $\mathcal{M}$ with states $s \in \mathcal{S}$, actions $a \in \mathcal{A}$, and time horizon $T$ \cite{puterman2014markov}. The robot does not have access to the reward function or transition dynamics of $\mathcal{M}$ but can cede control to a human supervisor, who executes some deterministic policy $\pisup: \mathcal{S} \to \mathcal{A}$. We refer to times when the robot is in control as \textit{autonomous mode} and those in which the supervisor is in control as \textit{supervisor mode}. 
We minimize a surrogate loss function $J(\pirob)$ to encourage the robot policy $\pirob: \mathcal{S} \to \mathcal{A}$ to match that of the supervisor ($\pisup$):
\begin{align}
    \label{eq:IL-objective}
    J(\pirob) = \sum_{t=1}^{T} \mathds{E}_{s_t \sim d^{\pirob}_t} \left[\mathcal{L}(\pirob(s_t),\pisup(s_t))\right],
\end{align}
where $\mathcal{L}(\pirob(s),\pisup(s))$ is an action discrepancy measure between $\pirob(s)$ and $\pisup(s)$ (e.g., the squared loss or 0-1 loss), and $d^{\pirob}_t$ is the marginal state distribution at timestep $t$ induced by executing $\pirob$ in MDP $\mathcal{M}$.

In interactive IL we require a meta-controller $\pimeta$ that determines whether to query the robot policy $\pirob$ or to query for an intervention from the human supervisor policy $\pisup$; importantly, $\pimeta$ consists of both (1) the high-level controller which decides whether to switch between $\pirob$ and $\pisup$ and (2) the low-level robot policy $\pirob$. A key objective in interactive IL is to minimize some notion of supervisor burden. To this end, let $\indicatorint(s_t; \pimeta)$ be an indicator which records whether a context switch between autonomous ($\pirob$) and supervisor ($\pisup$) modes occurs at state $s_t$ (either direction). Then, we define $C(\pimeta)$, the expected number of context switches in an episode under policy $\pimeta$, as follows: ${C(\pimeta) = \sum_{t=1}^{T} \mathds{E}_{s_t \sim d^{\pimeta}_t} \left[ \indicatorint(s_t; \pimeta) \right]}$, where $d^{\pimeta}_t$ is the marginal state distribution at timestep $t$ induced by executing the meta-controller $\pimeta$ in MDP $\mathcal{M}$. 
Similarly, let $\indicatorsup(s_t; \pimeta)$ indicate whether the system is in supervisor mode at state $s_t$. We then define $D(\pimeta)$, the expected number of supervisor actions in an episode for the policy $\pimeta$, as follows: ${D(\pimeta) = \sum_{t=1}^{T} \mathds{E}_{s_t \sim d^\pimeta_t} \left[ \indicatorsup(s_t; \pimeta) \right]}$.

We define \emph{supervisor burden} $B(\pimeta)$ as the expected time cost imposed on the human supervisor. This can be expressed as the sum of the expected total number of time units spent in context switching and the expected total number of time units in which the supervisor is actually engaged in performing interventions:
\begin{align}
    \label{eq:burden}
        B(\pimeta) = L \cdot C(\pimeta) + D(\pimeta),
\end{align}
where $L$ is context switch latency (Section~\ref{sec:introduction}) in time units, and each time unit is the time it takes for the supervisor to execute a single action. The learning objective is to find a policy $\pimeta$ that matches supervisor performance, $\pisup$, while limiting supervisor burden to lie within a threshold $\Gamma_{\rm b}$, set by the supervisor to an acceptable tolerance for a given task. To formalize this problem, we propose the following objective: 
\begin{align}
\begin{split}
    \label{eq:LazyDAgger-objective}
    \pimeta &= \argmin_{\pimeta' \in \Pi}\{ J(\pirob')
    \mid B(\pimeta') \leq \Gamma_{\rm b}\},
\end{split}
\end{align}
where $\Pi$ is the space of all meta-controllers, and $\pirob'$ is the low-level robot policy associated with meta-controller $\pimeta'$. 

\section{Preliminaries: SafeDAgger}
\label{sec:prelims}
We consider interactive IL in the context of the objective introduced in Equation~\eqref{eq:LazyDAgger-objective}: to maximize task reward while limiting supervisor burden. To do this, \algabbr builds on SafeDAgger~\cite{safe_dagger}, a state-of-the-art algorithm for interactive IL. SafeDAgger selects between autonomous mode and supervisor mode by training a binary action discrepancy classifier $\ffilt$ to discriminate between ``safe" states which have an action discrepancy below a threshold $\tausup$ (i.e., states with ${\mathcal{L}(\pirob(s), \pisup(s)) < \tausup}$) and ``unsafe" states (i.e. states with ${\mathcal{L}(\pirob(s), \pisup(s)) \geq \tausup}$). The classifier $\ffilt$ is a neural network with a sigmoid output layer (i.e., $\ffilt(s) \in [0, 1]$) that is trained to minimize binary cross-entropy (BCE) loss on the datapoints $(s_t,\pisup(s_t))$ sampled from a dataset $\mathcal{D}$ of trajectories collected from $\pisup$. This is written as follows:
\begin{align}
\label{eq:safe-loss}
    \begin{split}
    \lsafe(\pirob(s_t),\pisup(s_t), \ffilt) 
    = -\fstarfilt(\pirob(s_t),\pisup(s_t))\log \ffilt(s_t) \\
    -(1 - \fstarfilt(\pirob(s_t), \pisup(s_t)))\log (1 - \ffilt(s_t) ),
    \end{split}
\end{align}
where the training labels are given by $\fstarfilt(\pirob(s_t),\pisup(s_t)) = \mathbbm{1}\left\{\mathcal{L}(\pirob(s_t), \pisup(s_t)) \geq  \tausup\right\}$, and $\mathbbm{1}$ denotes the indicator function.
Thus, $\lsafe(\pirob(s_t),\pisup(s_t),\ffilt)$ penalizes incorrectly classifying a ``safe" state as ``unsafe" and vice versa. 

SafeDAgger executes the meta-policy $\pimeta$ which selects between $\pirob$ and $\pisup$ as follows:
\begin{align}
    \label{eq:safe_dagger_policy}
    \pimeta(s_t) =\begin{cases}
            \pirob(s_t) \text{ if } \ffilt(s_t) < 0.5 \\
            \pisup(s_t) \text{ otherwise},
          \end{cases}
\end{align}
where $f(s_t) < 0.5$ corresponds to a prediction that $\mathcal{L}(\pirob(s_t), \pisup(s_t)) < \tausup$, i.e., that $s_t$ is ``safe.'' Intuitively, SafeDAgger only solicits supervisor actions when $\ffilt$ predicts that the action discrepancy between $\pirob$ and $\pisup$ exceeds the safety threshold $\tausup$. Thus, SafeDAgger provides a mechanism for querying the supervisor for interventions only when necessary. In \algabbr, we utilize this same mechanism to query for interventions but enforce new properties once we enter these interventions to lengthen them and increase the diversity of states observed during the interventions.

\begin{figure}[b!]
\center
\includegraphics[width=\linewidth]{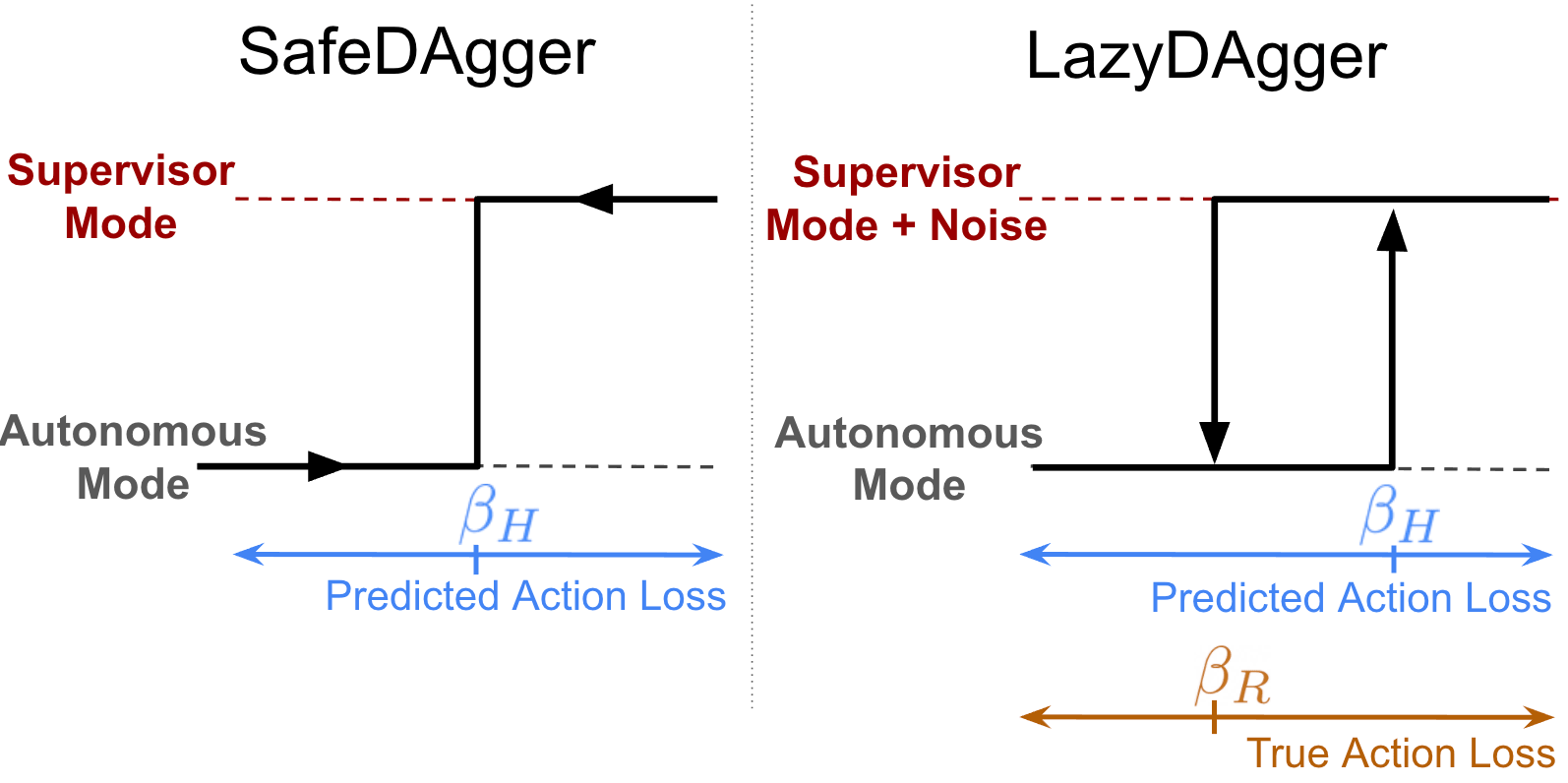}
\caption{
\textbf{\algname Switching Strategy:  }SafeDAgger switches between supervisor and autonomous mode if the predicted action discrepancy is above threshold $\tausup$. In contrast, \algname uses asymmetric switching criteria and switches to autonomous mode based on ground truth action discrepancy. The gap between $\tauauto$ and $\tausup$ defines a hysteresis band~\cite{hysteresis}.}
\label{fig:hyst}
\end{figure}
\section{\algname}
\label{sec:alg}

\begin{figure*}[htb!]
\center
\includegraphics[width=0.92\textwidth]{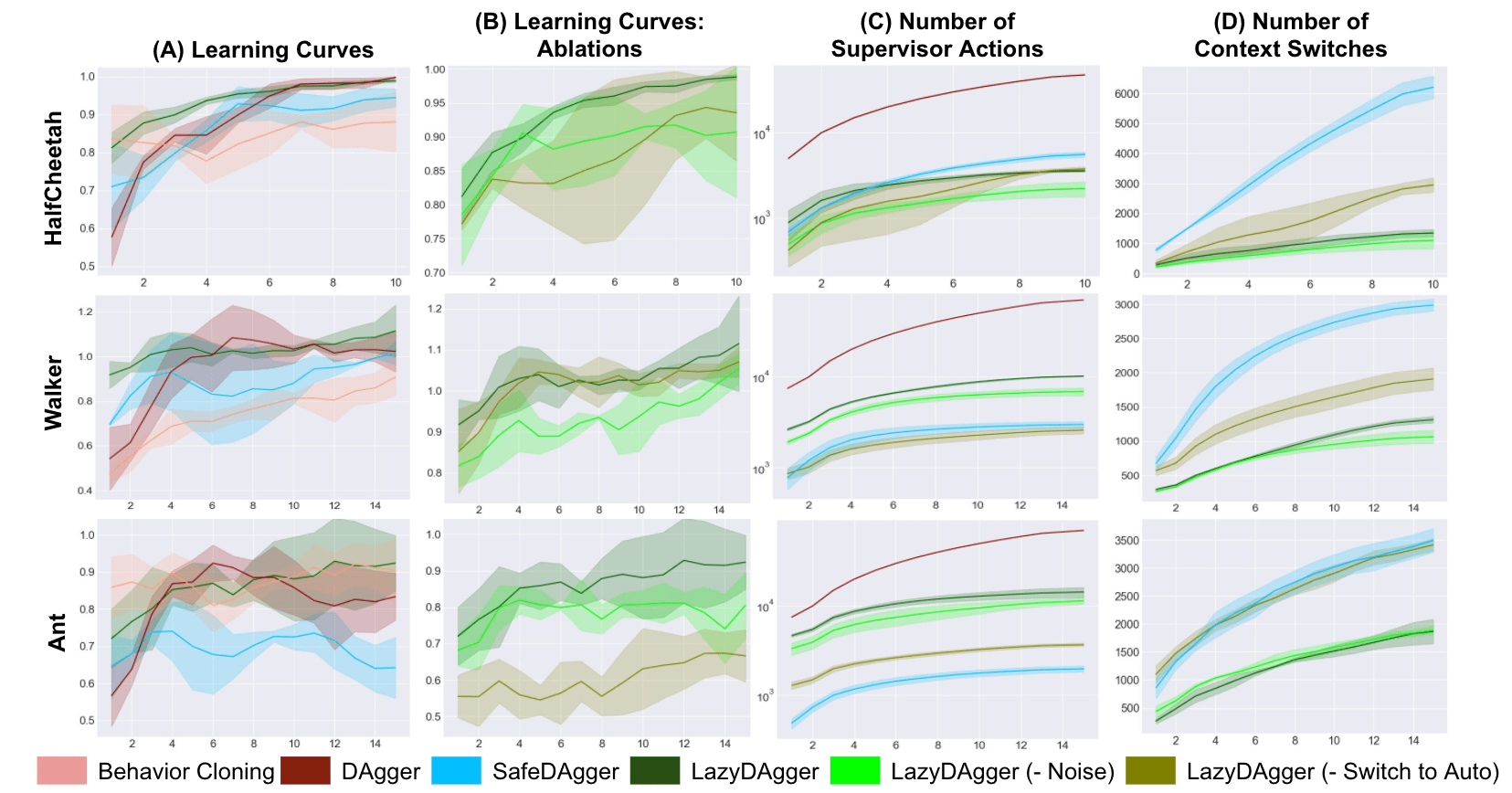}
\caption{
\textbf{MuJoCo Simulation Results: }We study task performance (A), ablations (B), online supervisor burden (C), and total bidirectional context switches (D) for \algname and baselines over 3 random seeds. For Columns (A)-(D), the x-axis for all plots shows the number of epochs over the training dataset, while the y-axes indicate normalized reward (A, B), counts of supervisor actions (C, log scale), and context switches (D) with shading for 1 standard deviation. We find that \algname outperforms all baselines and ablations, indicating that encouraging lengthy, noisy interventions improves performance. Additionally, \algname uses far fewer context switches than other baselines while requesting far fewer supervisor actions than DAgger.}
\label{fig:mujoco}
\end{figure*}

\begin{algorithm}[t]
\caption{\algname}
\label{alg:main}
\footnotesize
\begin{algorithmic}[1]
\Require Number of epochs $N$, time steps per epoch $T$, intervention thresholds $\tausup$, $\tauauto$, supervisor policy $\pisup$, noise $\sigma^2$ 
\State Collect ${\mathcal{D}}, \dsafe$ offline with supervisor policy $\pisup$
\State $\pirob \leftarrow \arg\min_{\pirob} \mathbb{E}_{(s_t, \pisup(s_t))\sim\mathcal{D}}\left[\cloningloss\right]$ \Comment{Eq.~\eqref{eq:IL-objective}}
\State $\ffilt \leftarrow \arg\min_{\ffilt} \mathbb{E}_{(s_t, \pisup(s_t))\sim\mathcal{D}\cup\dsafe} \left[\lsafe(\pirob(s_t),\pisup(s_t),\ffilt)\right]$ \Comment{Eq.~\eqref{eq:safe-loss}}
\For{$i \in \{1,\ldots N\}$}
    \State Initialize $s_0$, Mode $\gets$ Autonomous
    \For{$t \in \{1,\ldots T\}$}
        \State $a_t \sim \pirob(s_t)$
        \If{Mode = Supervisor or $\ffilt(s_t) \geq 0.5$}
            \State $a^H_t = \pisup(s_t)$
            \State $\mathcal{D} \leftarrow \mathcal{D} \cup \{(s_t, a_t^H)\}$
            \State Execute $\tilde{a}^H_t \sim \mathcal{N}(a^H_t,\sigma^2I)$ \label{lin:noise}
            \If {$\mathcal{L}(a_t, a^H_t) < \tauauto$}
                \State Mode $\gets$ Autonomous
            \Else
                \State Mode $\gets$ Supervisor
            \EndIf
        \Else
            \State Execute $a_{t}$
        \EndIf
    \EndFor
    \State $\pirob \leftarrow \arg\min_{\pirob} \mathbb{E}_{(s_t, \pisup(s_t))\sim\mathcal{D}}\left[\cloningloss\right]$
    \State $\ffilt \leftarrow \arg\min_{\ffilt} \mathbb{E}_{(s_t, \pisup(s_t))\sim \mathcal{D}\cup \dsafe}\left[\lsafe(\pirob(s_t),\pisup(s_t),\ffilt)\right]$
\EndFor
\end{algorithmic}
\end{algorithm}

We summarize \algabbr in Algorithm~\ref{alg:main}. In the initial phase (Lines 1-3), we train $\pirob$ and safety classifier $\ffilt$ on offline datasets collected from the supervisor policy $\pisup$. In the interactive learning phase (Lines 4-19), we evaluate and update the robot policy for $N$ epochs, ceding control to the supervisor when the robot predicts a high action discrepancy.

\subsection{Action Discrepancy Prediction}
\label{subsec:ac-disc}
SafeDAgger uses the classifier $\ffilt$ to select between $\pirob$ and $\pisup$ (Equation~\eqref{eq:safe_dagger_policy}). However, in practice, this often leads to frequent context switching (Figure~\ref{fig:mujoco}). 
To mitigate this, we make two observations. First, we can leverage that in supervisor mode, we directly observe the supervisor's actions. Thus, there is no need to use $\ffilt$, which may have approximation errors, to determine whether to remain in supervisor mode; instead, we can compute the ground-truth action discrepancy $\cloningloss$ exactly for any state $s_t$ visited in supervisor mode by comparing the supplied supervisor action $\pisup(s_t)$ with the action proposed by the robot policy $\pirob(s_t)$. In contrast, SafeDAgger uses $\ffilt$ to determine when to switch modes both in autonomous and supervisor mode, which can lead to very short interventions when $\ffilt$ prematurely predicts that the agent can match the supervisor's actions. Second, to ensure the robot has returned to the supervisor's distribution, the robot should only switch back to autonomous mode when the action discrepancy falls below a threshold $\tauauto$, where $\tauauto < \tausup$. 
As illustrated in Figure~\ref{fig:hyst}, \algname's asymmetric switching criteria create a hysteresis band, as is often utilized in control theory~\cite{hysteresis}. Motivated by Eq.~\eqref{eq:LazyDAgger-objective}, we adjust $\tausup$ to reduce context switches $C(\pimeta)$ and adjust $\tauauto$ as a function of $\tausup$ to increase intervention length. We hypothesize that redistributing the supervisor actions into fewer but longer sequences in this fashion both reduces burden on the supervisor and improves the quality of the online feedback for the robot. Details on setting these hyperparameter values in practice, the settings used in our experiments, and a hyperparameter sensitivity analysis are provided in the Appendix.


\subsection{Noise Injection}
\label{subsec:noise-inj}
If the safety classifier is querying for interventions at state $s_t$, then the robot either does not have much experience in the neighborhood of $s_t$ or has trouble matching the demonstrations at $s_t$. This motivates exploring novel states near $s_t$ so that the robot can receive maximal feedback on the correct behavior in areas of the state space where it predicts a large action discrepancy from the supervisor. Inspired by prior work that has identified noise injection as a useful tool for improving the performance of imitation learning algorithms (e.g.~\citet{DART} and~\citet{brown2019drex}), we diversify the set of states visited in supervisor mode by injecting isotropic Gaussian noise into the supervisor's actions, where the variance $\sigma^2$ is a scalar hyperparameter (Line \ref{lin:noise} in Algorithm~\ref{alg:main}).


\section{Experiments}
\label{sec:results}
\begin{figure*}[htb!]
\center
\includegraphics[width=0.85\textwidth]{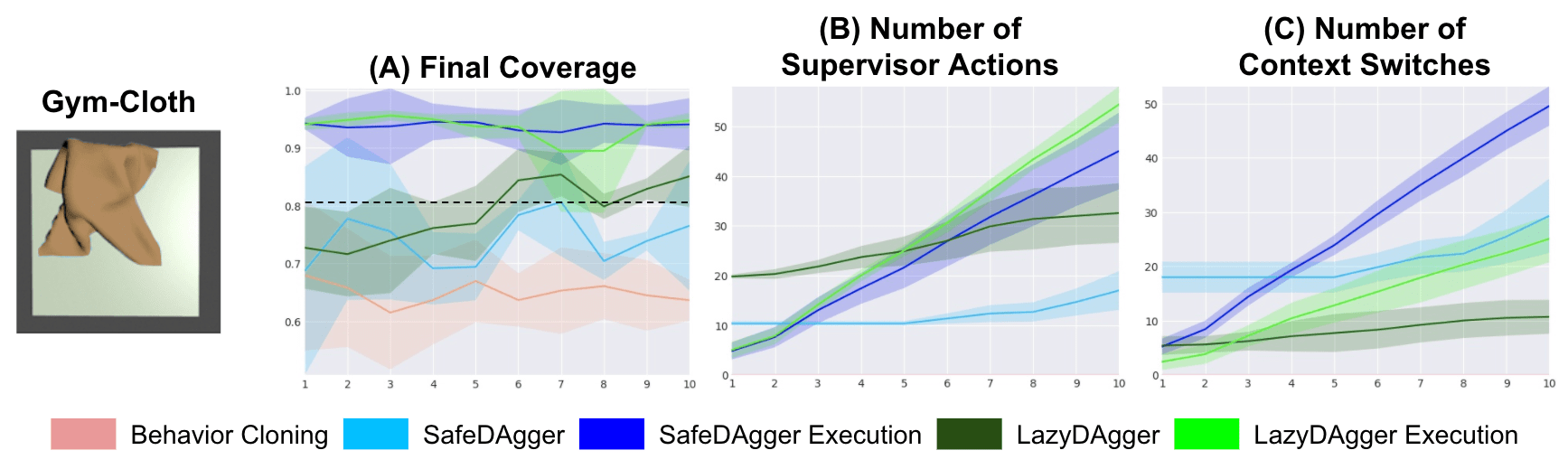}
\caption{
\textbf{Fabric Smoothing Simulation Results:} We study task performance measured by final fabric coverage (A), total supervisor actions (B), and total context switches (C) for \algname and baselines in the Gym-Cloth environment from~\cite{seita_fabrics_2020}. The horizontal dotted line shows the success threshold for fabric smoothing. \algname achieves higher final coverage than Behavior Cloning and SafeDAgger with fewer context switches than SafeDAgger but more supervisor actions. At execution time, we again observe that \algname achieves similar coverage as SafeDAgger but with fewer context switches.}
\label{fig:sim2sim}
\end{figure*}
We study whether \algname can (1) reduce supervisor burden while (2) achieving similar or superior task performance compared to prior algorithms. Implementation details are provided in the supplementary material. In all experiments, $\mathcal{L}$ measures Euclidean distance.  

\subsection{Simulation Experiments: MuJoCo Benchmarks}\label{ssec:mujoco-results}
\textbf{Environments: }We evaluate \algname and baselines on 3 continuous control environments from MuJoCo~\cite{mujoco}, a standard simulator for evaluating imitation and reinforcement learning algorithms. In particular, we evaluate on HalfCheetah-v2, Walker2D-v2 and Ant-v2.

\textbf{Metrics:}
\label{ssec:metrics}
For \algname and all baselines, we report learning curves which indicate how quickly they can make task progress in addition to metrics regarding the burden imposed on the supervisor. To study supervisor burden, we report the number of supervisor actions, the number of context switches, and the total supervisor burden (as defined in Eq.~\eqref{eq:burden}). Additionally, we define $L^* \geq 0$ to be the latency value such that for all $L > L^*$, \algname has a lower supervisor burden than SafeDAgger. We report this $L^*$ value, which we refer to as the \textit{cutoff latency}, for all experiments to precisely study the types of domains in which \algname is most applicable.

\textbf{Baselines:}
\label{ssec:baselines}
We compare \algname to Behavior Cloning~\cite{torabi2018behavioral}, DAgger~\cite{dagger}, and SafeDAgger~\cite{safe_dagger} in terms of the total supervisor burden and task performance. The Behavior Cloning and DAgger comparisons evaluate the utility of human interventions, while the comparison to SafeDAgger, another interactive IL algorithm, evaluates the impact of soliciting fewer but longer interventions.

\textbf{Experimental Setup: }For all MuJoCo environments, we use a reinforcement learning agent trained with TD3~\cite{td3} as an algorithmic supervisor. 
We begin all \algabbr, SafeDAgger, and DAgger experiments by pre-training the robot policy with Behavior Cloning on 4,000 state-action pairs for 5 epochs, and similarly report results for Behavior Cloning after the 5th epoch. To ensure a fair comparison, Behavior Cloning uses additional offline data equal to the average amount of online data seen by \algabbr during training. All results are averaged over 3 random seeds.

\textbf{Results: }In Figure~\ref{fig:mujoco}, we study the performance of \algabbr and baselines. After every epoch of training, we run the policy for 10 test rollouts \emph{where interventions are not allowed} and report the task reward on these rollouts in Figure~\ref{fig:mujoco}. Results suggest that \algabbr is able to match or outperform all baselines in terms of task performance across all simulation environments (Figure~\ref{fig:mujoco}A). Additionally, \algabbr requires far fewer context switches compared to SafeDAgger (Figure~\ref{fig:mujoco}D), while requesting a similar number of supervisor actions across domains (Figure~\ref{fig:mujoco}C): we observe a 79\%, 56\%, and 46\% reduction in context switches on the HalfCheetah, Walker2D, and Ant environments respectively. \algname and SafeDAgger both use an order of magnitude fewer supervisor actions than DAgger. While SafeDAgger requests much fewer supervisor actions than \algname in the Ant environment, this limited amount of supervision is insufficient to match the task performance of \algname or any of the baselines, suggesting that SafeDAgger may be terminating interventions prematurely. We study the total supervisor burden of SafeDAgger and \algname as defined in Equation~\eqref{eq:burden} and find that in HalfCheetah, Walker2D, and Ant, the cutoff latencies $L^*$ are 0.0, 4.3, and 7.6 respectively, i.e. \algname achieves lower supervisor burden in the HalfCheetah domain for any $L$ as well as lower burden in Walker2D and Ant for $L > 4.3$ and $L > 7.6$ respectively. The results suggest that \algname can reduce total supervisor burden compared to SafeDAgger even for modest latency values, but that SafeDAgger may be a better option for settings with extremely low latency.

\textbf{Ablations: } We study 2 key ablations for \algname in simulation: (1) returning to autonomous mode with $\ffilt(\cdot)$ rather than using the ground truth discrepancy (\algname (-Switch to Auto) in Figure~\ref{fig:mujoco}), and (2) removal of noise injection (\algname (-Noise)). \algabbr outperforms both ablations on all tasks, with the exception of ablation 1 on Walker2D, which performed similarly well. We also observe that \algabbr consistently requests more supervisor actions than either ablation. This aligns with the intuition that both using the ground truth action discrepancy to switch back to autonomous mode and injecting noise result in longer but more useful interventions that improve performance. 

\begin{figure*}[t]
\center
\includegraphics[width=0.90\textwidth]{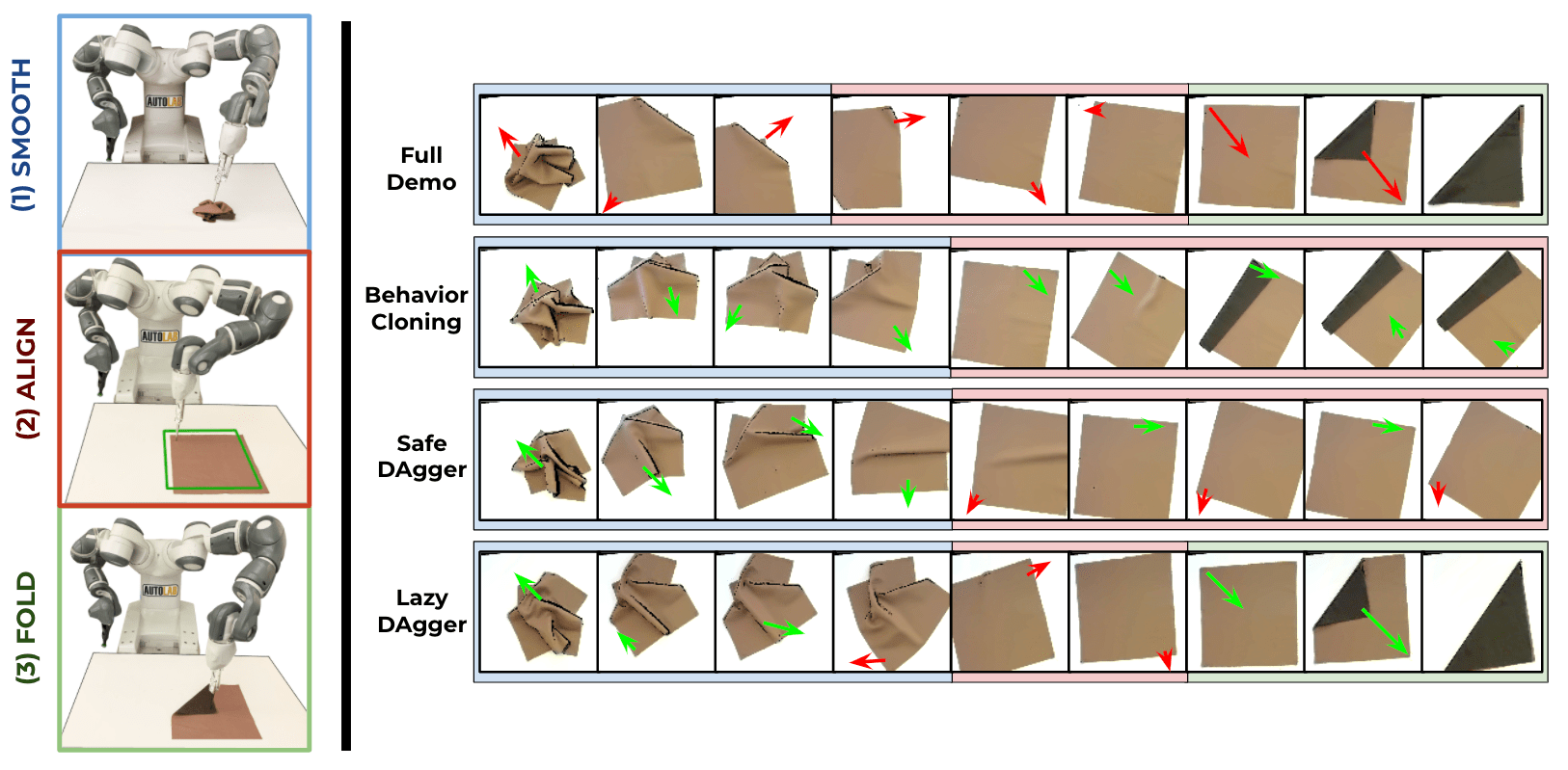}
\caption{
\textbf{Physical Fabric Manipulation Task: }\textit{Left: } We evaluate on a 3-stage fabric manipulation task consisting of smoothing a crumpled fabric, aligning the fabric so all corners are visible in the observations, and performing a triangular fold. \textit{Right: }  Rollouts of the fabric manipulation task, where each frame is a 100 $\times$ 100 $\times$ 3 overhead image. Human supervisor actions are denoted in red while autonomous robot actions are in green. Rollouts are shaded to indicate task progress: blue for smoothing, red for alignment, and green for folding. SafeDAgger ends human intervention prematurely, resulting in poor task performance and more context switches, while \algname switches back to robot control only when confident in task completion.}
\label{fig:phys-rollouts}
\end{figure*}

\begin{table*}[ht]
\centering
{
 \begin{tabular}{|l | c | c | c | c | c | c | c | c | c | c | c |} 
 \hline
 Algorithm & Task Successes & \multicolumn{3}{c|}{Task Progress} & Context Switches & Supervisor Actions & Robot Actions & \multicolumn{4}{c|}{Failure Modes} \\ 
 \hline
\multicolumn{2}{|c|}{} & (1) & (2) & (3) & \multicolumn{3}{c|}{} & A & B & C & D \\
 \hline
 Behavior Cloning & 0/10 & 6/10 & 0/10 & 0/10 & N/A & N/A & 119 & 2 & 1 & 7 & 0 \\
 \hline
 SD-Execution & 2/10 & 6/10 & 4/10 & 2/10 & 53 & \textbf{34} & 108 & 5 & 0 & 0 & 3 \\
 \hline
 LD-Execution & \textbf{8/10} & 10/10 & 10/10 & 8/10 & \textbf{21} & 43 & 47 & 0 & 0 & 0 & 2 \\
\hline
\end{tabular}}
\caption{\textbf{Physical Fabric Manipulation Experiments:} We evaluate \algname-Execution and baselines on a physical 3-stage fabric manipulation task and report the success rate and supervisor burden in terms of total supervisor actions and bidirectional context switches (summed across all 10 trials). Task Progress indicates how many trials completed each of the 3 stages: Smoothing, Aligning, and Folding. \algname-Execution achieves more successes with fewer context switches ($L^* = 0.28$). We observe the following failure modes (Table~\ref{tab:phys_results}): (A) action limit hit ($>$ 15 total actions), (B) fabric is more than 50\% out of bounds, (C) incorrect predicted pick point, and (D) the policy failed to request an intervention despite high ground truth action discrepancy.}
\label{tab:phys_results}
\end{table*}

\subsection{Fabric Smoothing in Simulation} \label{ssec:fabric_sim_results}
\textbf{Environment: }
We evaluate \algabbr on the fabric smoothing task from~\cite{seita_fabrics_2020} (shown in Figure~\ref{fig:sim2sim}) using the simulation environment from~\cite{seita_fabrics_2020}. The task requires smoothing an initially crumpled fabric and is challenging due to the infinite-dimensional state space and complex dynamics, motivating learning from human feedback. As in prior work~\cite{seita_fabrics_2020}, we utilize top-down $100 \times 100 \times 3$ RGB image observations of the workspace and use actions which consist of a 2D pick point and a 2D pull vector. See~\cite{seita_fabrics_2020} for further details on the fabric simulator.

\textbf{Experimental Setup: }
We train a fabric smoothing policy in simulation using DAgger under supervision from an analytic corner-pulling policy that leverages the simulator's state to identify fabric corners, iterate through them, and pull them towards the corners of the workspace~\cite{seita_fabrics_2020}. We transfer the resulting policy for a 16$\times$16 grid of fabric into a new simulation environment with altered fabric dynamics (i.e. lower spring constant, altered fabric colors, and a higher-fidelity 25$\times$25 discretization) and evaluate \algname and baselines on how rapidly they can adapt the initial policy to the new domain. As in~\cite{seita_fabrics_2020}, we terminate rollouts when we exceed 10 time steps, 92\% coverage, or have moved the fabric more than 20\% out of bounds. We evaluate performance based on a coverage metric, which measures the percentage of the background plane that the fabric covers (fully smooth corresponds to a coverage of 100).

\textbf{Results: }
We report results for the fabric smoothing simulation experiments in Figure~\ref{fig:sim2sim}. Figure~\ref{fig:sim2sim}~(A) shows the performance of the SafeDAgger and \algname policies during learning. To generate this plot we periodically evaluated each policy on \emph{test} rollouts without interventions. Figure~\ref{fig:sim2sim}~(B) and (C) show the number of supervisor actions and context switches required during learning; \algname performs fewer context switches than SafeDAgger but requires more supervisor actions as the interventions are longer. Results suggest that the cutoff latency (as defined in Section~\ref{ssec:metrics}) is $L^*=1.5$ for fabric smoothing. Despite fewer context switches, \algname achieves comparable performance to SafeDAgger, suggesting that \algname can learn complex, high-dimensional robotic control policies while reducing the number of hand-offs to a supervisor. We also evaluate \algname-Execution and SafeDAgger-Execution, in which interventions are allowed but the policy is no longer updated (see Section~\ref{ssec:physresults}). We see that in this case, \algname achieves similar final coverage as SafeDAgger with significantly fewer context switches. 

\subsection{Physical Fabric Manipulation Experiments}\label{ssec:physresults}
\textbf{Environment: }
In physical experiments, we evaluate on a multi-stage fabric manipulation task with an ABB YuMi robot and a human supervisor (Figure~\ref{fig:phys-rollouts}). Starting from a crumpled initial fabric state, the task consists of 3 stages: (1) fully smooth the fabric, (2) align the fabric corners with a tight crop of the workspace, and (3) fold the fabric into a triangular fold. Stage (2) in particular requires high precision, motivating human interventions. As in the fabric simulation experiments, we use top-down $100 \times 100 \times 3$ RGB image observations of the workspace and have 4D actions consisting of a pick point and pull vector. The actions are converted to workspace coordinates with a standard calibration procedure and analytically mapped to the nearest point on the fabric. Human supervisor actions are provided through a point-and-click interface for specifying pick-and-place actions. See the supplement for further details.

\textbf{Experimental Setup: }
Here we study how interventions can be leveraged to improve the final task performance even at \textit{execution time}, in which policies are no longer being updated. We collect 20 offline task demonstrations and train an initial policy with behavior cloning. To prevent overfitting to a small amount of real data, we use standard data augmentation techniques such as rotating, scaling, changing brightness, and adding noise to create 10 times as many training examples. We then evaluate the behavior cloning agent (Behavior Cloning) and agents which use the SafeDAgger and \algname intervention criteria but do not update the policy with new experience or inject noise (SafeDAgger-Execution and \algname-Execution respectively). We terminate rollouts if the fabric has successfully reached the goal state of the final stage (i.e. forms a perfect or near-perfect dark brown right triangle as in~\citet{fabric_vsf}; see Figure~\ref{fig:phys-rollouts}), more than 50\% of the fabric mask is out of view in the current observation, the predicted pick point misses the fabric mask by approximately 50\% of the plane or more, or 15 total actions have been executed (either autonomous or supervisor).

\textbf{Results: }
We perform 10 physical trials of each technique. In Table~\ref{tab:phys_results}, we report both the overall task success rate and success rates for each of the three stages of the task: (1) Smoothing, (2) Alignment, and (3) Folding. We also report the total number of context switches, supervisor actions, and autonomous robot actions summed across all 10 trials for each algorithm (Behavior Cloning, SafeDAgger-Execution, \algname-Execution). In Figure~\ref{fig:phys-rollouts} we provide representative rollouts for each algorithm. Results suggest that Behavior Cloning is insufficient for successfully completing the alignment stage with the required level of precision. SafeDAgger-Execution does not improve the task success rate significantly due to its inability to collect interventions long enough to navigate bottleneck regions in the task (Figure~\ref{fig:phys-rollouts}). \algname-Execution, however, achieves a much higher success rate than SafeDAgger-Execution and Behavior Cloning with far fewer context switches than SafeDAgger-Execution: \algname-Execution requests 2.1 context switches on average per trial (i.e. 1.05 interventions) as opposed to 5.3 switches (i.e. 2.65 interventions). \algname-Execution trials also make far more task progress than the baselines, as all 10 trials reach the folding stage. \algname-Execution does request more supervisor actions than SafeDAgger-Execution, as in the simulation environments. \algname-Execution also requests more supervisor actions relative to the total amount of actions due to the more conservative switching criteria and the fact that successful episodes are shorter than unsuccessful episodes on average. Nevertheless, results suggest that for this task, \algname-Execution reduces supervisor burden for any $L > L^* = 0.28$, a very low cutoff latency that includes all settings in which a context switch is at least as time-consuming as an individual action (i.e. $L \geq 1$).

In experiments, we find that SafeDAgger-Execution's short interventions lead to many instances of Failure Mode A (see Table~\ref{tab:phys_results}), as the policy is making task progress, but not quickly enough to perform the task. We observe that Failure Mode C is often due to the fabric reaching a highly irregular configuration that is not within the training data distribution, making it difficult for the robot policy to make progress. We find that SafeDAgger and \algname experience Failure Mode D at a similar rate as they use the same criteria to solicit interventions (but different termination criteria). However, we find that all of \algname's failures are due to Failure Mode D, while SafeDAgger also fails in Mode A due to premature termination of interventions.
\section{Discussion and Future Work}
\label{sec:discussion}
We propose context switching between robot and human control as a metric for supervisor burden in interactive imitation learning and present \algname, an algorithm which can be used to efficiently learn tasks while reducing this switching. We evaluate \algname on 3 continuous control benchmark environments in MuJoCo, a fabric smoothing environment in simulation, and a fabric manipulation task with an ABB YuMi robot and find that \algname is able to improve task performance while reducing context switching between the learner and robot by up to 79\% over SafeDAgger. 
In subsequent work, we investigate more intervention criteria and apply robot-gated interventions to controlling a fleet of robots, where context switching can negatively impact task throughput.

\section{Acknowledgments}
\footnotesize
This research was performed at the AUTOLAB at UC Berkeley in affiliation with the Berkeley AI Research (BAIR) Lab, and the CITRIS ``People and Robots" (CPAR) Initiative. This research was supported in part by the Scalable Collaborative Human-Robot Learning (SCHooL) Project, NSF National Robotics Initiative Award 1734633. The authors were supported in part by donations from Google, Siemens, 
Toyota Research Institute, Autodesk, Honda, Intel, and Hewlett-Packard and by 
equipment grants from PhotoNeo, Nvidia, and Intuitive Surgical. Any opinions, findings, and conclusions or recommendations expressed in this material are those of the author(s) and do not necessarily reflect the views of the sponsors. 
We thank colleagues Lawrence Chen, Jennifer Grannen, and Vincent Lim for providing helpful feedback and suggestions.

\renewcommand{\bibfont}{\footnotesize}

\printbibliography
\clearpage
\normalsize
\section{Appendix}
\label{sec:appendix}
Here we provide further details on our MuJoCo experiments, hyperparameter sensitivity, simulated fabric experiments, and physical fabric experiments.

\subsection{MuJoCo}\label{ssec:mujoco-appdx}
As stated in the main text, we evaluate on the HalfCheetah-v2, Walker2D-v2, and Ant-v2 environments. To train the algorithmic supervisor, we utilize the TD3 implementation from OpenAI SpinningUp (\url{https://spinningup.openai.com/en/latest/}) with default hyperparameters and run for 100, 200, and 500 epochs respectively. The expert policies obtain rewards of 5330.78 $\pm$ 117.65, 3492.08 $\pm$ 1110.31, and 4492.88 $\pm$ 1580.42, respectively. Note that the experts for Walker2D and Ant have high variance, resulting in higher variance for the corresponding learning curves in Figure~\ref{fig:mujoco}. We provide the state space dimensionality $|S|$, action space dimensionality $|A|$, and \algabbr hyperparameters (see Algorithm~\ref{alg:main}) for each environment in Table~\ref{tab:mujoco-params}. The $\tausup$ value in the table is multiplied with the maximum possible action discrepancy $||a_{\rm high} - a_{\rm low}||_2^2$ to become the threshold for training $\ffilt(\cdot)$. In MuJoCo environments, $a_{\rm high} = \vec 1$ and $a_{\rm low} = -\vec 1$. The $\tausup$ value used for SafeDAgger in all experiments is chosen by the method provided in the paper introducing SafeDAgger~\cite{safe_dagger}: the threshold at which roughly 20\% of the initial offline dataset is classified as ``unsafe."

For \algabbr and all baselines, the actor policy $\pirob(\cdot)$ is a neural network with 2 hidden layers with 256 neurons each, rectified linear unit (ReLU) activation, and hyperbolic tangent output activation. For \algabbr and SafeDAgger, the discrepancy classifier $\ffilt(\cdot)$ is a neural network with 2 hidden layers with 128 neurons each, ReLU activation, and sigmoid output activation. We take 2,000 gradient steps per epoch and optimize with Adam and learning rate 1e-3 for both neural networks. To collect $\mathcal{D}$ and $\dsafe$ in Algorithm~\ref{alg:main} and SafeDAgger, we randomly partition our dataset of 4,000 state-action pairs into 70\% (2,800 state-action pairs) for $\mathcal{D}$ and 30\% (1,200 state-action pairs) for $\dsafe$.

\begin{table}[!htbp]
\centering
\resizebox{\columnwidth}{!}{
 \begin{tabular}{l | c | c | c | c | c | c | c } 
 Environment & $|S|$ & $|A|$ & $N$ & $T$ & $\tausup$ & $\tauauto$ & $\sigma^2$ \\ 
 \hline
 HalfCheetah & 16 & 7 & 10 & 5000 & 5e-3 & $\tausup$ / 10 & 0.30 \\
 Walker2D & 16 & 7 & 15 & 5000 & 5e-3 & $\tausup$ / 10 & 0.10 \\
 Ant & 111 & 8 & 15 & 5000 & 5e-3 & $\tausup$ / 2 & 0.05 \\
\end{tabular}}
\caption{\textbf{MuJoCo Hyperparameters:} $|S|$ and $|A|$ are aspects of the Gym environments while the other values are hyperparameters of LazyDAgger (Algorithm~\ref{alg:main}). Note that $T$ and $\tausup$ are the same across all environments, and that $\tauauto$ is a function of $\tausup$.}
\label{tab:mujoco-params}
\end{table}

\subsection{\algabbr Switching Thresholds}
As described in Section~\ref{subsec:ac-disc}, the main \algabbr hyperparameters are the safety thresholds for switching to supervisor control ($\tausup$) and returning to autonomous control ($\tauauto$). To tune these hyperparameters in practice, we initialize $\tausup$ and $\tauauto$ with the method in~\citet{safe_dagger}; again, this sets the safety threshold such that approximately 20\% of the initial dataset is unsafe. We then tune $\tausup$ higher to balance reducing the frequency of switching to the supervisor with allowing enough supervision for high policy performance. Finally we set $\tauauto$ as a multiple of $\tausup$, starting from $\tauauto = \tausup$ and tuning downward to balance improving the performance and increasing intervention length with keeping the total number of actions moderate. Note that since these parameters are not automatically set, we must re-run experiments for each change of parameter values. Since this tuning results in unnecessary supervisor burden, eliminating or mitigating this requirement is an important direction for future work.

\begin{figure}[t]
\center
\includegraphics[width=0.45\textwidth]{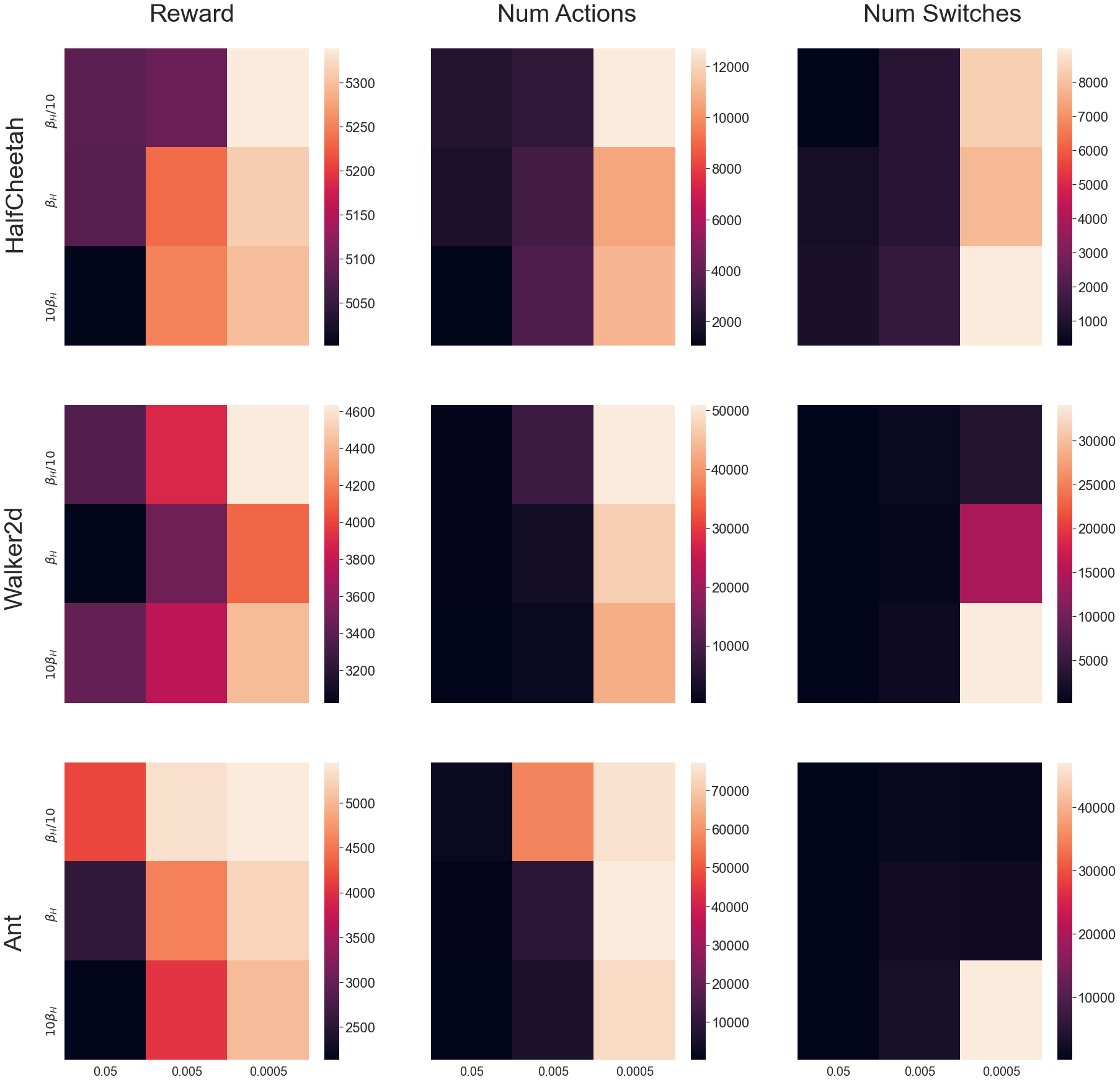}
\caption{
\algabbr $\tauauto$ and $\tausup$ sensitivity heatmaps across the 3 MuJoCo environments.  The x-axis denotes $\tausup$ and the y-axis denotes $\tauauto$.  Note that $\tauauto$ is a function of $\tausup$.  Each of the 3 environments was run 9 times with the different settings of $\tauauto$ and $\tausup$. As in Figure~\ref{fig:mujoco} we plot test reward, number of supervisor actions, and number of context switches.}
\label{fig:tau_heat_map}
\end{figure}

To analyze sensitivity to $\tauauto$ and $\tausup$, we plot the results of a grid search over parameter values on each of the MuJoCo environments in Figure~\ref{fig:tau_heat_map}. Note that a lighter color in the heatmap is more desirable for reward while a darker color is more desirable for actions and switches. We see that the supervisor burden in terms of actions and context switches is not very sensitive to the threshold as we increase $\tausup$ but jumps significantly for the very low setting ($\tausup = 5\times 10^{-4}$) as a large amount of data points are classified as unsafe. Similarly, we see that reward is relatively stable (note the small heatmap range for HalfCheetah) as we decrease $\tausup$ but can suffer for high values, as interventions are not requested frequently enough. Reward and supervisor burden are not as sensitive to $\tauauto$ but follow the same trends we expect, with higher reward and burden as $\tauauto$ decreases.

\subsection{Fabric Smoothing in Simulation}

\subsubsection{Fabric Simulator}\label{ssec:fabricsim}

More information about the fabric simulator can be found in~\citet{seita_fabrics_2020}, but we review the salient details here. The fabric is modeled as a mass-spring system with a $n \times n$ square grid of point masses.
Self-collision is implemented by applying a repulsive force between points that are sufficiently close together. Blender (\url{https://blender.org/}) is used to render the fabric in $100 \times 100 \times 3$ RGB image observations. See Figure~\ref{fig:sim2sim} for an example observation. The actions are 4D vectors consisting of a pick point $(x, y) \in [-1,1]^2$ and a place point $(\Delta x, \Delta y) \in [-1,1]^2$, where $(x,y) = (-1,-1)$ corresponds to the bottom left corner of the plane while $(\Delta x, \Delta y)$ is multiplied by 2 to allow crossing the entire plane. In simulation, we initialize the fabric with coverage $41.1 \pm 3.4\%$ in the hardest (Tier 3) state distribution in~\cite{seita_fabrics_2020} and end episodes if we exceed 10 time steps, cover at least 92\% of the plane, are at least 20\% out of bounds, or have exceeded a tearing threshold in one of the springs. We use the same algorithmic supervisor as~\cite{seita_fabrics_2020}, which repeatedly picks the coordinates of the corner furthest from its goal position and pulls toward this goal position. To facilitate transfer to the real world, we use the domain randomization techniques in~\cite{seita_fabrics_2020} to vary the following parameters: 
\begin{itemize}
    \item Fabric RGB values uniformly between (0, 0, 128) and (115, 179, 255), centered around blue.
    \item Background plane RGB values uniformly between (102, 102, 102) and (153, 153, 153).
    \item RGB gamma correction uniformly between 0.7 and 1.3.
    \item Camera position ($x, y, z$) as $(0.5+\delta_1, 0.5+\delta_2, 1.45+\delta_3)$ meters, where each $\delta_i$ is sampled from $\mathcal{N}(0, 0.04)$.
    \item Camera rotation with Euler angles sampled from $\mathcal{N}(0, 90^{\degree} )$.
    \item Random noise at each pixel uniformly between -15 and 15.
\end{itemize}
For consistency, we use the same domain randomization in our sim-to-sim (``simulator to simulator") fabric smoothing experiments in Section \ref{ssec:fabric_sim_results}.

\subsubsection{Actor Policy and Discrepancy Classifier}\label{ssec:actorpol}
The actor policy is a convolutional neural network with the same architecture as~\cite{seita_fabrics_2020}, i.e. four convolutional layers with 32 3x3 filters followed by four fully connected layers. The parameters, ignoring biases for simplicity, are:

\footnotesize
\begin{verbatim}
policy/convnet/c1   864 params (3, 3, 3, 32)
policy/convnet/c2   9216 params (3, 3, 32, 32)
policy/convnet/c3   9216 params (3, 3, 32, 32)
policy/convnet/c4   9216 params (3, 3, 32, 32)
policy/fcnet/fc1    3276800 params (12800, 256)
policy/fcnet/fc2    65536 params (256, 256)
policy/fcnet/fc3    65536 params (256, 256)
policy/fcnet/fc4    1024 params (256, 4)
Total model parameters: 3.44 million
\end{verbatim}
\normalsize

The discrepancy classifier reuses the actor's convolutional layers by taking a forward pass through them. We do not backpropagate gradients through these layers when training the classifier, but rather fix these parameters after training the actor policy. The rest of the classifier network has three fully connected layers with the following parameters:

\footnotesize
\begin{verbatim}
policy/fcnet/fc1    3276800 params (12800, 256)
policy/fcnet/fc2    65536 params (256, 256)
policy/fcnet/fc3    1024 params (256, 4)
Total model parameters: 3.34 million
\end{verbatim}
\normalsize

\subsubsection{Training}
\begin{figure}[t]
\center
\includegraphics[width=0.45\textwidth]{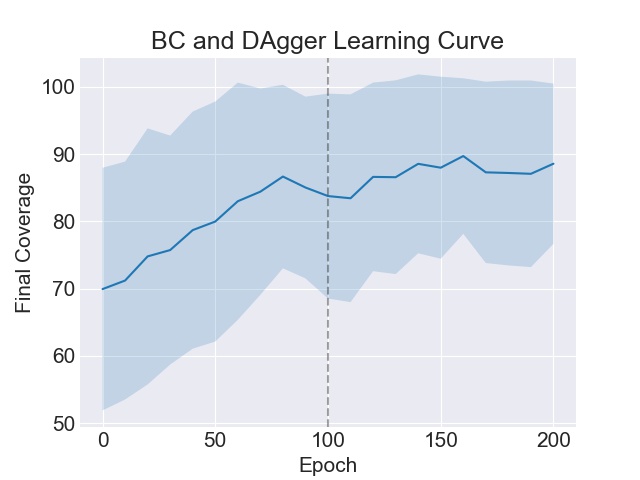}
\caption{
Behavior Cloning and DAgger performance across 10 test episodes evaluated every 10 epochs. Shading indicates 1 standard deviation. The first 100 epochs (left half) are Behavior Cloning epochs and the second 100 (right half) are DAgger epochs.}
\label{fig:daggersim}
\end{figure}

Due to the large amount of data required to train fabric smoothing policies, we pretrain the \textit{actor policy} (not the discrepancy classifier) in simulation. The learned policy is then fine-tuned to the new environment while the discrepancy classifier is trained from scratch. Since the algorithmic supervisor can be queried cheaply, we pretrain with DAgger as in~\cite{seita_fabrics_2020}. To further accelerate training, we parallelize environment interaction across 20 CPUs, and before DAgger iterations we pretrain with 100 epochs of Behavior Cloning on the dataset of 20,232 state-action pairs available at~\cite{seita_fabrics_2020}'s project website. Additional training hyperparameters are given in Table~\ref{tab:dagger-params} and the learning curve is given in Figure~\ref{fig:daggersim}. 

\begin{table}[!htbp]
\centering
{
 \begin{tabular}{l r} 
Hyperparameter & Value \\
\hline
BC Epochs & 100 \\
DAgger Epochs & 100 \\
Parallel Environments & 20 \\
Gradient Steps per Epoch & 240 \\
Env Steps per Env per DAgger Epoch & 20 \\
Batch Size & 128 \\
Replay Buffer Size & 5e4 \\
Learning Rate & 1e-4 \\
L2 Regularization & 1e-5 \\
\end{tabular}}
\caption{\textbf{DAgger Hyperparameters.} After Behavior Cloning, each epoch of DAgger (1) runs the current policy and collects expert labels for 20 time steps in each of 20 parallel environments and then (2) takes 240 gradient steps on minibatches of size 128 sampled from the replay buffer.}
\label{tab:dagger-params}
\end{table}

\subsubsection{Experiments}
In sim-to-sim experiments, the initial policy is trained on a 16x16 grid of fabric in a range of colors centered around blue with a spring constant of $k=10,000$. We then adapt this policy to a new simulator with different physics parameters and an altered visual appearance. Specifically, in the new simulation environment, the fabric is a higher fidelity 25x25 grid with a lower spring constant of $k=2,000$ and a color of (R, G, B) = (204, 51, 204) (i.e. pink), which is outside the range of colors produced by domain randomization (Section~\ref{ssec:fabricsim}). Hyperparameters are given in Table~\ref{tab:s2s-params}.

\begin{table}[!htbp]
\centering
{
 \begin{tabular}{l r} 
Hyperparameter & Value \\
\hline
$N$ & 10 \\
$T$ & 20 \\
$\tausup$ & 0.001 \\
$\tauauto$ & $\tausup$  \\
$\sigma^2$ & 0.05 \\
Initial $|\mathcal{D}|$ & 1050 \\
Initial $|\dsafe|$ & 450 \\
Batch Size & 50 \\
Gradient Steps per Epoch & 200 \\
$\pi$ Learning Rate & 1e-4 \\
$\ffilt$ Learning Rate & 1e-3 \\
L2 Regularization & 1e-5 \\
\end{tabular}}
\caption{Hyperparameters for sim-to-sim fabric smoothing experiments, where the first 5 rows are \algabbr hyperparameters in Algorithm~\ref{alg:main}. Initial dataset sizes and batch size are in terms of images \textit{after} data augmentation, i.e. scaled up by a factor of 15 (see Section~\ref{ssec:dataaug}). Note that the offline data is split 70\%/30\% as in Section~\ref{ssec:mujoco-appdx}.}
\label{tab:s2s-params}
\end{table}

\subsection{Fabric Manipulation with the ABB YuMi}
\subsubsection{Experimental Setup}
We manipulate a brown 10" by 10" square piece of fabric with a single parallel jaw gripper as shown in Figure~\ref{fig:teaser}. The gripper is equipped with reverse tweezers for more precise picking of deformable materials. Neural network architecture is consistent with Section~\ref{ssec:actorpol} for both actor and safety classifier. We correct pick points that nearly miss the fabric by mapping to the nearest point on the mask of the fabric, which we segment from the images by color. To convert neural network actions to robot grasps, we run a standard camera calibration procedure and perform top-down grasps at a fixed depth. By controlling the width of the tweezers via the applied force on the gripper, we can reliably pick only the top layer of the fabric at a given pick point. We provide \algabbr-Execution hyperparameters in Table~\ref{tab:real-params}. 

\subsubsection{Image Processing Pipeline}
In the simulator, the fabric is smoothed against a light background plane with the same size as the fully smoothed fabric (see Figure~\ref{fig:sim2sim}). Since the physical workspace is far larger than the fabric, we process each RGB image of the workspace by (1) taking a square crop, (2) rescaling to 100 $\times$ 100, and (3) denoising the image. Essentially we define a square crop of the workspace as the region to smooth and align against, and assume that the fabric starts in this region. These processed images are the observations that fill the replay buffer and are passed to the neural networks. 

\subsubsection{User Interface}
\begin{figure}[t]
\center
\includegraphics[width=0.3\textwidth]{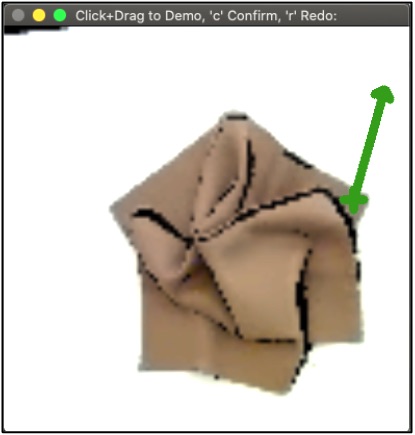}
\caption{
The user interface for human interventions. The current observation of the fabric state from the robot's perspective is displayed, with an overlaid green arrow indicating the action the human has just specified.}
\label{fig:ui}
\end{figure}

When the system solicits human intervention, an interactive user interface displays a scaled-up version of the current observation. The human is able to click and drag on the image to provide a pick point and pull vector, respectively. The interface captures the input as pixel locations and analytically converts it to the action space of the environment (i.e. $a \in [-1,1]^4$) for the robot to execute. See Figure~\ref{fig:ui} for a screen capture of the user interface.

\subsubsection{Data Augmentation}\label{ssec:dataaug}
To prevent overfitting to the small amount of real data, before adding each state-action pair to the replay buffer, we make 10 copies of it with the following data augmentation procedure, with transformations applied in a random order:
\begin{itemize}
    \item Change contrast to 85-115\% of the original value.
    \item Change brightness to 90-110\% of the original value.
    \item Change saturation to 95-105\% of the original value.
    \item Add values uniformly between -10 and 10 to each channel of each pixel.
    \item Apply a Gaussian blur with $\sigma$ between 0 and 0.6.
    \item Add Gaussian noise with $\sigma$ between 0 and 3.
    \item With probability 0.8, apply an affine transform that (1) scales each axis independently to 98-102\% of its original size, (2) translates each axis independently by a value between -2\% and 2\%, and (3) rotates by a value between -5 and 5 degrees.
\end{itemize}

\begin{table}[!htbp]
\centering
{
 \begin{tabular}{l r} 
Hyperparameter & Value \\
\hline
$\tausup$ & 0.004 \\
$\tauauto$ & $\tausup$ \\
$|\mathcal{D}|$ & 875 \\
$|\dsafe|$ & 375 \\
Batch Size & 50 \\
Gradient Steps per Epoch & 125 \\
$\pi$ Learning Rate & 1e-4 \\
$\ffilt$ Learning Rate & 1e-3 \\
L2 Regularization & 1e-5 \\
\end{tabular}}
\caption{Hyperparameters for physical fabric experiments provided in the same format as Table~\ref{tab:s2s-params}. Since this is at execution time, $N$, $T$ and $\sigma^2$ hyperparameters do not apply.}
\label{tab:real-params}
\end{table}
\end{document}